%% file: root.tex
\documentclass[letterpaper, 10 pt, conference]{ieeeconf}  

\IEEEoverridecommandlockouts                              

\usepackage{graphicx}
\usepackage{romannum}
\usepackage{amsmath}
\usepackage{diagbox}
\usepackage{makecell}
\usepackage{hyperref}
\usepackage{amssymb}
\usepackage{bm}
\usepackage{subcaption}
\usepackage{mathtools}
\usepackage{booktabs}
\usepackage{bm}
\usepackage{xcolor}
\usepackage{soul}
\usepackage{cite}
\usepackage{tabularray}

\title{\LARGE \bf
Fully Distributed Cooperative Multi-agent Underwater Obstacle Avoidance}
\author{Kanzhong Yao
\and Ognjen Marjanovic \and Simon Watson
\thanks{This work was supported by Chinese Scholarship Council-University of Manchester joint programme. The authors acknowledge the support provided by EPSRC (Hot Robotics: EP/T011491/1) and the Robotics and AI Collaboration (RAICo).}
\thanks{All authors are with Manchester Centre for Robotics and AI, Department of Electrical and Electronic Engineering, University of Manchester, UK. Please
direct correspondence to {\tt\small kanzhong.yao@postgrad.manchester.ac.uk}}%
}
\begin{document}
\maketitle
\thispagestyle{empty}
\pagestyle{empty}

\begin{abstract}
\input{Subsections/0_abstract}
\end{abstract}
\section{Introduction}
\input{Subsections/1_introduction}
\section{Related Work}

\input{Subsections/2_related_work}
\section{ Dog Walking Paradigm Principles}
\input{Subsections/3_dog_walking}
\section{Cooperative Underwater Obstacle Avoidance}
\input{Subsections/4_proposed_solution}
\section{Experiments}
\input{Subsections/5_experiments}
\section{Conclusion \& Future Work}
\input{Subsections/6_conclusion}
\bibliographystyle{IEEEtran}
\bibliography{reference.bib}
\end{document}

%% file: Subsections/0_abstract.tex
Navigation in cluttered underwater environments is challenging, especially when there are constraints on communication and self-localisation. Part of the fully distributed underwater navigation problem has been resolved by introducing multi-agent robot teams \cite{Yao2024}, however when the environment becomes cluttered, the problem remains unresolved. In this paper, we first studied the connection between everyday activity of dog walking and the cooperative underwater obstacle avoidance problem. Inspired by this analogy, we propose a novel dog walking paradigm and implement it in a multi-agent underwater system.  Simulations were conducted across various scenarios, with performance benchmarked against traditional methods utilising Image-Based Visual Servoing in a multi-agent setup. Results indicate that our dog walking-inspired paradigm significantly enhances cooperative behavior among agents and outperforms the existing approach in navigating through obstacles. 



%% file: Subsections/1_introduction.tex
Autonomous exploration and inspection in hazardous environments represent one of the primary applications of robotic systems, with confined underwater spaces being typical examples of such challenging environments \cite{Watson2020}. These scenarios are prevalent across various domains, including aquaculture, nuclear storage ponds, and cave exploration.

Confined underwater spaces are characterised by several generic challenges: constrained communications \cite{Qu2016}, a lack of continuous features for self-localisation \cite{Xanthidis_2020}, and the absence of auxiliary facilities \cite{Watson2020}. The deployment of multi-robot teams has been proposed to mitigate some of these issues, offering enhanced operational capabilities and redundancy \cite{YAO2023}. However, when the environment becomes cluttered, an additional challenge will be brought to the multi-agent robots working in a team: How to achieve collision-free underwater motion whilst keeping in a certain formation of the team? 

Traditional approaches to addressing this challenge typically depend on centralised control systems, where a networked infrastructure orchestrates coordination \cite{Bai2022}, or partially centralised mechanisms that necessitate minimal levels of communication \cite{Hu2021}. On the other hand, fully distributed solutions often presuppose access to predetermined obstacle data \cite{Cai2020} or rely heavily on self-localisation capabilities to ensure collision-free navigation within formations \cite{Chen2017a}.

In scenarios of extreme environments, such as the confined underwater spaces previously discussed, collision-free formations that consider \textbf{no communication}, and \textbf{no underwater self-localisation} are sometimes necessary. Unfortunately, the problem under such conditions has been explored minimally. Part of our goal with this paper is to define problems under similar contexts into a generalised set, allowing for a more systematic study within the community, therefore more generalisable solutions can be proposed.   
  \begin{figure}[tpb]
      \centering
       \includegraphics[width=0.45\textwidth]{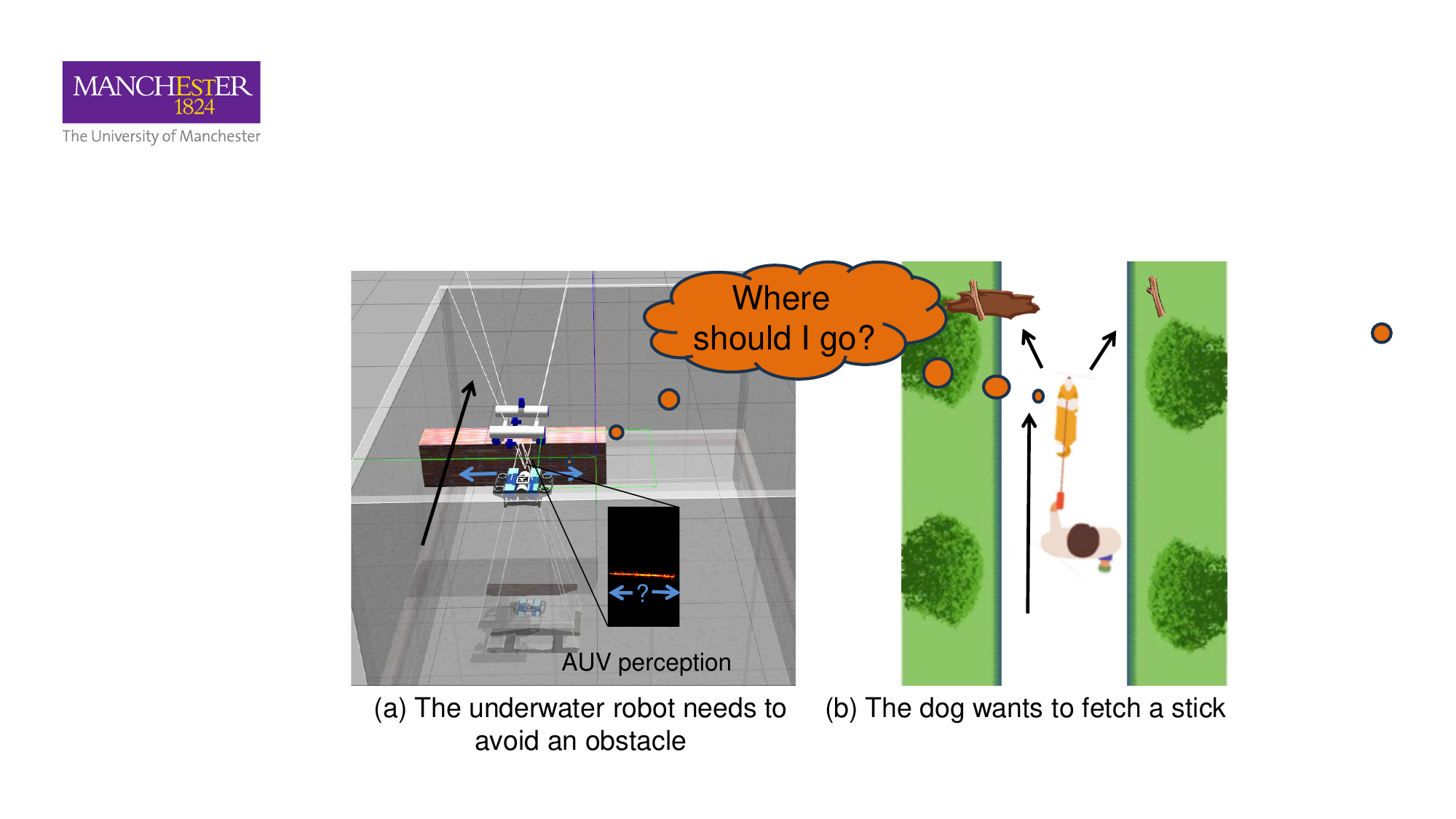}
      \caption{Fully distributed cooperative underwater obstacle avoidance and a common dog walking scenario. 
      }
      \vspace{-15pt}
      \label{fig:intro}
  \end{figure}
\section{Problem Statement \& Contribution}
Consider a scenario involving a cluttered, confined underwater environment where underwater localisation and communication are highly constrained. An underwater robot is dependent on guidance from a surface robot, faced with the additional complication that underwater obstacles can only be detected by the underwater robot within a restricted range. This situation necessitates that the underwater robot not only maintain formation with the guiding surface robot but also navigate around obstacles adeptly to ensure safe, collision-free movement. Meanwhile, the surface robot guides the underwater towards a target location, and also needs to actively adjust its motion to keep the formation with the underwater robot \cite{Yao2024}, as shown in Fig.~\ref{fig:intro}a.

Taking the inspiration from the analogy of walking a dog, in this work, we define a novel ``dog walking" paradigm and adapt it to solve the described problem. Given a multi-agent system, one ``follower" agent with limited sensing capabilities is guided towards a target position by a ``leader" agent. Analogous to walking a dog, the follower-dog must perform additional motions to fetch a stick, while maintaining formation with the leader-walker, in the absence of direct communication between them, as depicted in Fig.~\ref{fig:intro}b. The contributions of this work are:
\begin{itemize}
    \item We define a novel ``dog walking'' paradigm, which can be applied to solve the fully distributed cooperative obstacle avoidance problem. To the best of the authors' knowledge, it is the first in the literature.
    \item A preliminary solution with the new paradigm to the described problem was implemented, in the context of an underwater multi-agent system. Corresponding simulation has been conducted to demonstrate its efficacy, benchmarked against previous work \cite{Yao2024}.
\end{itemize}


%% file: Subsections/2_related_work.tex
Over the years, the cooperative obstacle avoidance problem in constrained environments has gained notable attention. In this section, we present an overview across aerial, ground, and underwater domains, as there are relatively few attempts in the underwater domain.

Under a constrained environment, where data transmission between agents is not allowed, the problem becomes distributed. For example, a deep reinforcement learning strategy was employed for non-communicative path optimisation to facilitate multi-agent collision avoidance in \cite{Chen2017a}. This was extended by approaches focusing on decentralised connectivity maintenance using reinforcement learning for Unmanned Aerial Vehicles (UAVs) \cite{Huang2022a}. Such strategies have similarly been applied to multi-Unmanned Ground Vehicle (UGV) \cite{Li2022c} and multi-Autonomous Underwater Vehicle (AUV) systems \cite{Fang2022}, showcasing the potential for fully distributed navigation in cluttered environments. A commonality among these studies is an underlying assumption regarding the agents' awareness of their positioning in a global frame. These works presuppose either the provision of agent and obstacle positions as ground truth data \cite{Chen2017a,Fang2022,Huang2022a} or the availability of high-precision perception methods like LiDAR and depth cameras \cite{Li2022c}. 

In practical applications, where perfect perception is often unattainable, researchers have begun to explore solutions accommodating imperfect measurements. For instance, Panagou et al. \cite{Panagou_2014} introduced a visibility maintenance framework designed to support cooperative leader-follower navigation through obstacle-laden environments with constrained visual input. Similarly, Dergachev \cite{Dergachev2021} developed a novel solver aimed at cooperative collision avoidance that accounts for limited sensory input. 

In the underwater domain, characterised by inherently difficult communication and self-localisation challenges, literature addressing these specific issues remains scarce. A notable attempt to navigate AUVs in cluttered environments involved the adaptation of an enhanced Trajopt algorithm \cite{Schulman2014}, as demonstrated by Xanthidis et al. \cite{Xanthidis_2020}. Their experimental setup required the placement of artificial objects at the pool's bottom to aid in robot localisation. This reliance on artificial markers underscores a significant limitation: in genuine real-world scenarios, devoid of such artificial aids, the efficacy of these systems is questionable.

To summarise, if we move slightly further, eliminating both the communication and self-localisation prerequisites, we will be closer to the reality of operating within extreme environments. However, no current approaches can solve the problem described in Section \MakeUppercase{\romannumeral 2} under such settings. 

    

%% file: Subsections/3_dog_walking.tex
  \begin{figure*}[tpb]
      \centering
       \includegraphics[width=0.8\textwidth]{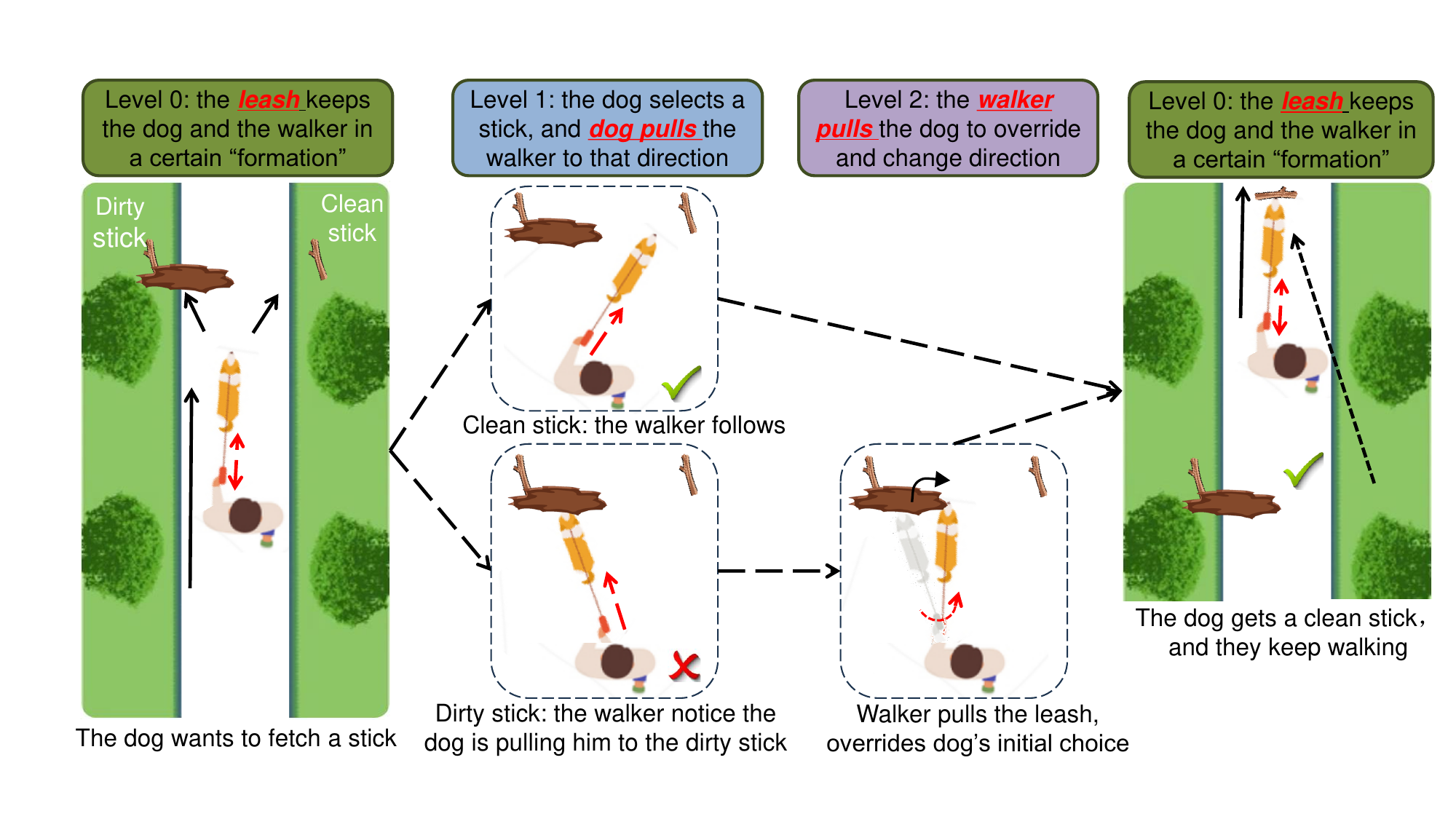}
      \caption{Dog-Walking Paradigm Illustration. The three-tiered connection hierarchy is color-coded: Level 0 in green, Level 1 in blue, and Level 2 in purple, the higher the level, the higher the priority. Implicit communication is highlighted in red, with red dashed arrows representing the non-verbal cues between the dog and the walker.}
      \vspace{-15pt}
      \label{fig:dog_walking}
  \end{figure*}

Dog walking, a common activity in daily life, exemplifies a dynamic interaction between two agents bound by a unique form of connection and cooperation. In this analogy, the `leader agent'—the person holding the leash—possesses a broader perspective of the environment and dominates the decision-making in critical situations such as avoiding hazardous areas. Conversely, the `follower agent'—the dog—operates with a more localised perception of its surroundings and may exhibit spontaneous behaviors, such as attempting to fetch a stick. These agents are linked by a leash, serving as the medium for implicit communication, notably through variations in tension, which can be seen as pulling or resisting movements.

A scenario (Fig.~\ref{fig:dog_walking}) where a dog expresses the desire to fetch a stick, presented with two options: one adjacent to the muddy puddle and another situated on the grass. The objective is to ensure the dog's retrieval of the clean stick. This process leverages two principal characteristics:
\subsubsection{Hierarchical Implicit Communication}
The communication hierarchy is demonstrated in Fig.~\ref{fig:dog_walking} and can be dissected into three distinct levels. Level 0 represents the initial state where the dog and the walker are united by a flexible leash, sharing a common goal of walking home. At Level 1, the dog communicates its intent to procure a specific stick, employing an implicit signaling method manifested through the act of leash pulling. Progressing to Level 2, the walker, perceives with a broader awareness of the environment, intervenes when the dog targets the dirty stick near the muddy puddle and initiates pulling, the walker will override the pulling from the dog also via implicit communication - pulling back. This corrective action persists until the dog adjusts its preference, ultimately steering clear of the muddy area. Similar implicit communication has been studied in \cite{Berlinger2021} \cite{Hwang2023}, but not in a hierarchical manner. 

\subsubsection{Formation Constraint}
The leash serves the physical connection that maintains the proximity of the dog and walker, embodying the concept of a formation constraint. This constraint is integral to the leader-follower system, ensuring the agents remain within a bounded distance of each other.
 


%% file: Subsections/4_proposed_solution.tex
In this section, a system model for the problem described in Section \MakeUppercase{\romannumeral 2} was formulated, including the fundamental model and the extended formulation components with the dog walking paradigm. Within such a system, BlueROV2 was utilised as the underwater robot while MallARD \cite{Groves2019} was the surface robot.

\subsection{System Model}
Both the BlueROV2 and MallARD are omnidirectional in terms of maneuverability, and the kinematic model for each robot can be expressed obtained in previous work \cite{Groves2020,YAO2023}. The dynamic model for both robots is approximated by the Fossen equation \cite{Fossen1995}. Therefore, the system model $J$ can be expressed as follows: 
\begin{subequations}\label{eq:system_model}
\begin{align}
    &{\dot{ \bm x}}^S = \bm{J}^S(\prescript{\mathcal{B}}{}{\bm{u}}^S)\\
    &{\dot{ \bm x}}^U = \bm{J}^U(\prescript{\mathcal{B}}{}{\bm{u}}^U)
\end{align}
\end{subequations}
the superscriptions $S$ and $U$ indicate the corresponding notations for ASV and AUV, where $\bm{x} = [x,y,z,\phi, \theta, \psi]^\top$, and $x,y,z$ denote the robot's position while the angles $\phi, \theta, \psi$ represent its orientation around $x$, $y$, and $z$, respectively. The robots' control inputs in the body frame $\prescript{\mathcal{B}}{}{\bm{u}}$ are represented by $\prescript{\mathcal{B}}{}{\bm{u}}^S =  [\prescript{\mathcal{B}}{}{u}_{x}^S, \prescript{\mathcal{B}}{}{u}_{y}^S, \prescript{\mathcal{B}}{}{u}_{\psi}^S]^\top$ and $\prescript{\mathcal{B}}{}{\bm{u}}^U =  [\prescript{\mathcal{B}}{}{u}_{x}^U, \prescript{\mathcal{B}}{}{u}_{y}^U, \prescript{\mathcal{B}}{}{u}_{z}^U, \prescript{\mathcal{B}}{}{u}_{\phi}^U, \prescript{\mathcal{B}}{}{u}_{\theta}^U, \prescript{\mathcal{B}}{}{u}_{\psi}^U]^\top$.

In practice, the ASV needs to determine the target position while keeping in formation with the AUV. Meanwhile, the AUV must also keep in formation with the ASV while avoiding underwater obstacles. Therefore, the control inputs can be written as: 
\begin{subequations}\label{eq:control_inputs_sub}
\begin{align}
    &\prescript{\mathcal{B}}{}{\bm{u}}^S = K_P \cdot \prescript{\mathcal{P}}{}{\bm{u}}^S + K_V \cdot \prescript{\mathcal{V}}{}{\bm{u}}^S\\
    &\prescript{\mathcal{B}}{}{\bm{u}}^U =   \prescript{\mathcal{O}}{}{\bm{u}}^U+  \prescript{\mathcal{V}}{}{\bm{u}}^U+ \prescript{\mathcal{P}}{}{\bm{u}}^U
\end{align}
\end{subequations}
where $\prescript{\mathcal{P}}{}{\bm{u}}^S$ is calculated by the global planner and PD controller developed in \cite{Groves2019}, which will send the ASV to a target position during an exploration mission. Control inputs $\prescript{\mathcal{V}}{}{\bm{u}}^S$ and $\prescript{\mathcal{V}}{}{\bm{u}}^U$ are generated by Virtual Elastic Tether (VET) mentioned in \cite{Yao2024}, which keeps the robots in each others' LoS using a mutual visual-based connection. $\prescript{\mathcal{O}}{}{\bm{u}}^U$ represents the control inputs for the AUV to avoid an obstacle. $\prescript{\mathcal{P}}{}{\bm{u}}^U$ controls the AUV's depth and roll, pitch angle to a target value, which in our implementation was set to $depth=-1.5$~m, $roll = 0$, and $pitch = 0$. At the initial state, weights meet $K_P=K_V $ and $\prescript{\mathcal{P}}{}{\bm{u}}^S_{\text{max}} = \prescript{\mathcal{V}}{}{\bm{u}}^S_{\text{max}}$, which means balance weighted are assigned in terms of the ASV following target position and ASV following AUV.

\subsection{Onboard Perceptions}
The formation of the robots is achieved via Image Based Visual serving (IBVS), and the obstacle detection is based on image sonar on the AUV, meanwhile, the ASV is equipped with a 2D LiDAR for self-localisation \cite{Groves2019} and wall detection, as depicted in Fig.~\ref{fig:perception}.
\subsubsection{Visual Based Formation}
\begin{figure}[tpb]
      \centering
       \includegraphics[width=0.4\textwidth]{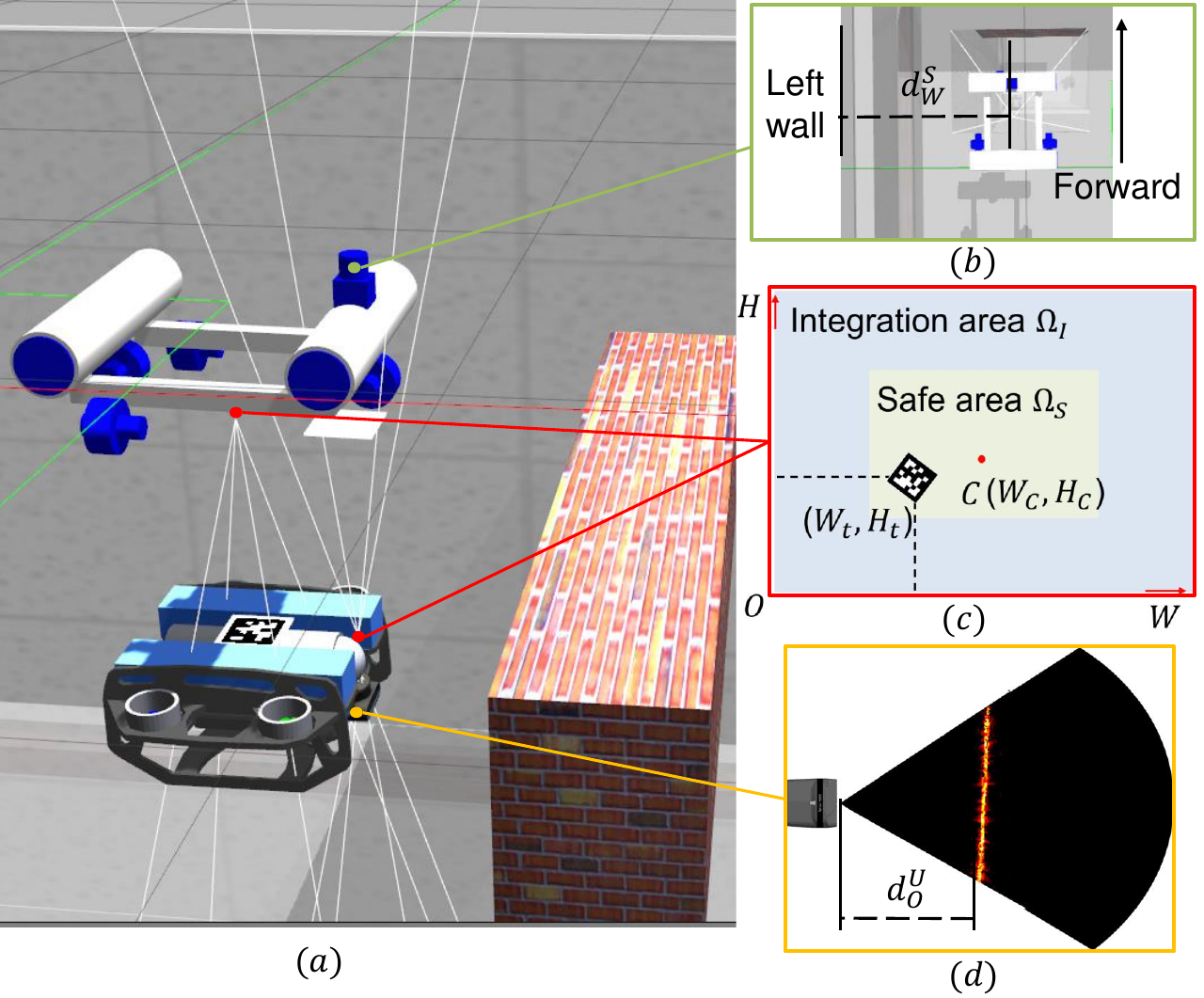}
      \caption{Illustration of onboard perception: a), the overall setup within a simulated tank environment; b), the ASV's mechanism for detecting the distance \({d}^S_W\) to the nearest wall, with \({d}^S_W < 0\) indicating a left-side wall and \({d}^S_W > 0\) for a right-side wall; c), the shared visual perspective of both robots, with the tag \((W_t, H_t)\) denoting the position of the robot in the image view at time $t$; d), the use of image sonar by the AUV for obstacle detection.}
      \vspace{-15pt}
      \label{fig:perception}
  \end{figure}

Based on the IBVS \cite{YAO2023} leader-follower formation and VET \cite{Yao2024}, the ASV and AUV in our system inherit such visual-based ``connection'':
\begin{subequations}\label{eq:IBVS}
\begin{align}
    &\prescript{\mathcal{V}}{}{\bm{u}}^S =\left\{ \begin{array}{lr}
       \Xi^S(W^U_t, H^U_t)  & (W^U_t,H^U_t)\in \Omega_S\\
     \Xi^S_{\text{max}} & (W^U_t,H^U_t)\in \Omega_I
    \end{array} \right.\\
    &\prescript{\mathcal{V}}{}{\bm{u}}^U =\left\{ \begin{array}{lr}
       \Xi^U(W^S_t, H^S_t)  & (W^S_t,H^S_t)\in \Omega_S\\
     \Xi^U_{\text{max}} & (W^S_t,H^S_t)\in \Omega_I
    \end{array} \right.
\end{align}
\end{subequations}
where $\Xi$ is the corresponding controller that functions on both robots to bring each other from $(W_t,H_t)$ to the centre $C(W_C,H_C)$ of its image view, as depicted in Fig.~\ref{fig:perception} (c). $W_C$ and $H_C$ are determined by the camera resolution. The integration area $\Omega_I$ and safe area $\Omega_S$ indicate the outer and inner area of the image view, as depicted in Fig.~\ref{fig:perception} (c).   

\subsubsection{Image Sonar Based Obstacle Perception}
In terms of obstacle detection, a multibeam image sonar was utilised on the AUV, as depicted in Fig.~\ref{fig:perception} (d), which in practice was handled by a physics-based plugin \cite{Choi2021}. The sonar was placed in the front bottom of BlueROV2, facing forward, which gives the online reading of the obstacle in front of the robot. By binarization and clustering the image, and mapping to the effective range of the sonar, we can obtain the distance from the AUV to the obstacle $d^U_O$.

\subsubsection{LiDAR Based Wall Perception}
The 2D LiDAR mounted on the ASV serves two primary functions: 1) it provides data for surface self-localization, and 2) it detects the proximity of the nearest wall to the ASV,  with \({d}^S_W < 0\) indicating a left-side wall and \({d}^S_W > 0\) indicating proximity to a right-side wall, as shown in Fig.~\ref{fig:perception} (b). 

\subsection{Dog Walking Paradigm Implementation: Principles}

  \begin{figure*}[tpb]
      \centering
       \includegraphics[width=0.9\textwidth]{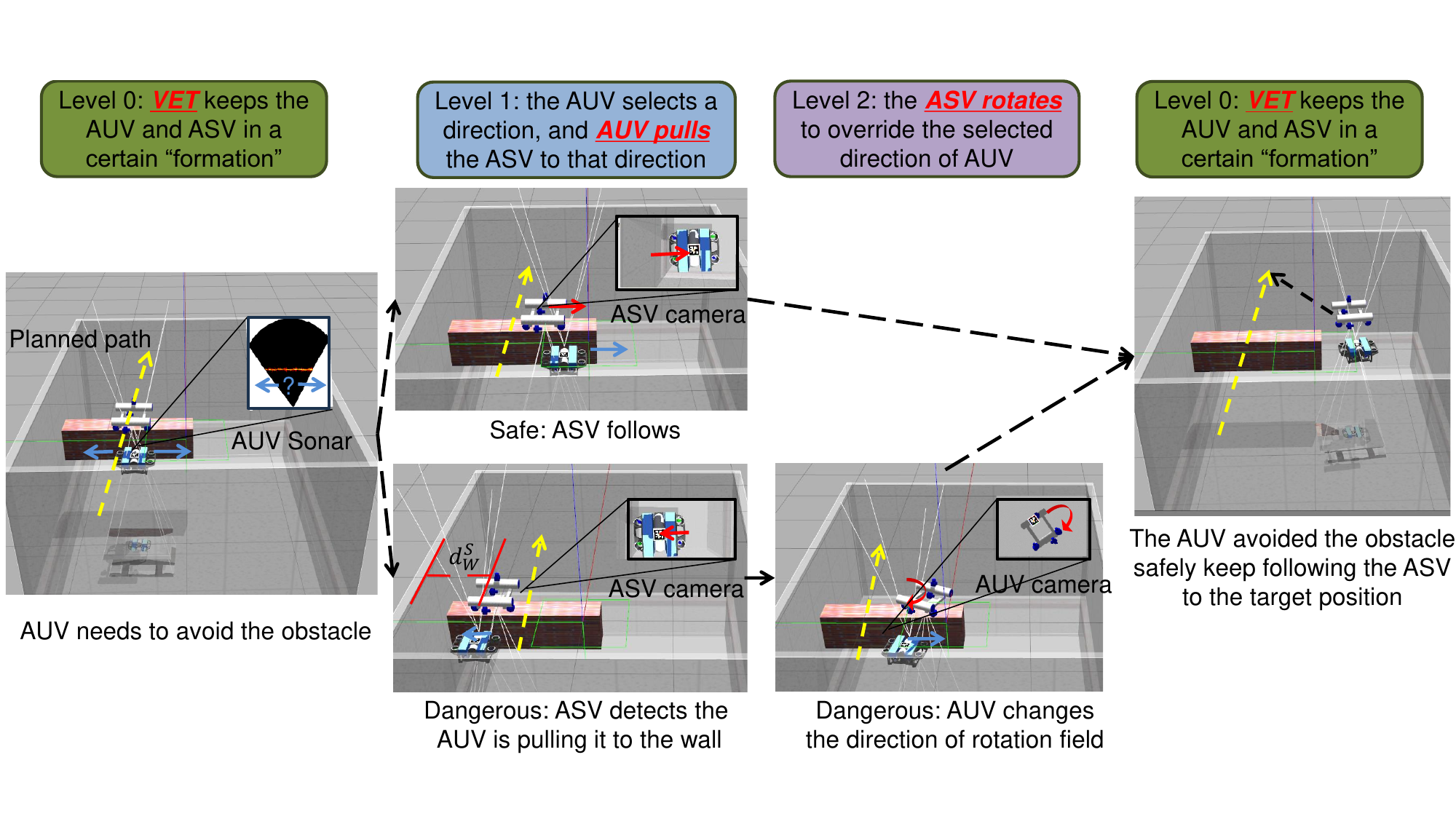}
      \caption{Cooperative underwater obstacle avoidance with dog walking paradigm. Similar to Fig.~\ref{fig:dog_walking}, the three-tiered (Level 0-2) connection hierarchy is color-coded in green, blue, and purple, respectively. Implicit communication is highlighted in red, with red arrows representing the interaction between the ASV and the AUV.}
      \vspace{-15pt}
      \label{fig:dog_walking_practice}
  \end{figure*}

To apply the dog walking paradigm in the context of underwater cooperative obstacle avoidance, the following basic principles can be designed:   
\subsubsection{Hierarchical Implicit Communication}
Similar to the dog walking scenario, the communication hierarchy is demonstrated in Fig.~\ref{fig:dog_walking_practice} and can be dissected into three levels. \textbf{Level 0} represents that the ASV and AUV are ``connected'' via a visual-based connection, as per (\ref{eq:IBVS}a-b).

At \textbf{Level 1}, the AUV detects the obstacle within the sonar range. When the distance to the obstacle meets $d^U_O \leq d^U_S$, where $d^U_S$ represents the safe distance to the obstacle, collision avoidance action must be taken: 
\begin{equation}\label{eq:level_1_0}
   \left[ \begin{array}{c}
        \prescript{\mathcal{O}}{}{{u}}^U_x  \\[5pt]
        \prescript{\mathcal{O}}{}{{u}}^U_y
   \end{array} \right]  = \left[ \begin{array}{c} \text{min} \{ \frac{\alpha}{d^U_O},\Xi^U_{\text{max}|x} \} \\[5pt] 
         (-1)^\lambda \cdot \text{min} \{ \frac{\alpha}{d^U_O}, \Xi^U_{\text{max}|y}\} 
    \end{array}\right]  
\end{equation}
where $\alpha$ is a consistent parameter for calculating the repulsive and rotational force from the obstacle. Other subparts of $\prescript{\mathcal{O}}{}{\bm{u}}^U$ are equal to $0$, and $\lambda$ denotes the directional parameter. Therefore, from (\ref{eq:control_inputs_sub}b), (\ref{eq:IBVS}b), and (\ref{eq:level_1_0}), we can obtain that when an obstacle is getting close, the AUV will move into $\Omega_I$ and eventually $\prescript{\mathcal{O}}{}{{u}}^U = \prescript{\mathcal{V}}{}{{u}}^U$, the robot will be stuck in an equilibrium position. To avoid that, the ASV must notice such equilibrium and take action. A weighting mechanism was designed to dynamically set $K_p$ and $K_V$ in (\ref{eq:control_inputs_sub}a): 
\begin{equation}\label{eq:level_1_1}
    \left\{ \begin{array}{lr}
       K_P = K_V  & \int_{t - t_o}^{t}\,k(t)\,dt< t_o \\[5pt]
     K_P = \beta \cdot K_V & \int_{t - t_o}^{t}\,k(t)\,dt \geq t_o
    \end{array} \right.\\
\end{equation}
above equation determines the relationship between $K_P$ and $K_V$ at time $t$, where the interval time $t_o$ and weighting coefficient $\beta$ are consistent parameter, $\beta$ meets $0<\beta<1$, and $k(t)$ can be written as follows: 
\begin{equation}\label{eq:level_1_2}
   k(t)= \left\{ \begin{array}{lr}
       0  & (W^U_t,H^U_t)\in \Omega_S \\[5pt]
     1 & (W^U_t,H^U_t)\in \Omega_I
    \end{array} \right.,
\end{equation}
(\ref{eq:level_1_1}) and (\ref{eq:level_1_2}) denote that when the AUV stays in the integration area $\Omega_I$ over a specified time $t_o$ due to the equilibrium, the control inputs of the formation in (\ref{eq:control_inputs_sub}) will be assigned higher weights than target following, therefore the ASV will be pulled towards the AUV to let the AUV avoid the obstacle until the AUV gets back in the safe area $\Omega_S$ again, as depicted at the top of the second stage in Fig.~\ref{fig:dog_walking_practice}.

However, as the AUV does not perceive the information where the wall is, it can not evaluate different direction selections. Conversely, as the ASV perceives the location of the wall, when the AUV selects a dangerous direction to avoid the obstacle, as depicted at the bottom of the second stage in Fig.~\ref{fig:dog_walking_practice}, the ASV must correct such direction selection via \textbf{Level 2} implicit communication mechanism.
  \begin{table}[tbp]
  \centering
  \label{tab:algorithm1}
  \begin{tabular}{rp{5.8cm}}   
  \toprule    
  {\textbf{Algorithm \MakeUppercase{\romannumeral 1}:}}&{\textbf{Level 2 indicator - ASV} }                                              \\[3pt]
    \toprule  
      \textbf{Input:} & $(W^U_t,H^U_t)$, $d^S_W$, $d^S_S$, $t_1$ and $\prescript{B}{}{u^S_\psi}$  \\[3pt]
        \textbf{1:}&  \textbf{while} $|d^S_W| \leq d^S_S \quad \text{and} \quad (W^U_t,H^U_t)\in \Omega_I$ \textbf{Do:} \\[3pt]
        \textbf{2:}&   \quad   $ \prescript{\mathcal{B}}{}{u^S_\psi}= \prescript{\mathcal{B}}{}{u^S_\psi} +  \frac{d^S_W}{|d^S_W|} \cdot \prescript{\mathcal{C}}{}{u^S_\psi} $       \\[3pt]
        \textbf{3:}&  \qquad $\mu = \frac{d^S_W}{|d^S_W|}$ \\[3pt]
        \textbf{4:}&  \qquad  \text{Sleep for time} \, $t_1$  \\[3pt]
        \textbf{Output:} &$\prescript{B}{}{u^S_\psi}$, $\mu$  \\
             \bottomrule 
  \end{tabular} 
        \vspace{-15pt}
  \end{table}
  
In the context of dog walking, when a walker perceives that the dog is pulling towards an undesirable object, such as a dirty stick, corrective action is taken by sharply yanking the leash. This creates a significant pulling force, overriding the dog's stick choice. Analogously, when the ASV detects that the AUV is pulling it towards a potentially hazardous proximity to a wall, the ASV initiates a rapid rotational maneuver away from the nearest wall. This maneuver is detailed in Algorithm~\MakeUppercase{\romannumeral 1}. Here, $\prescript{\mathcal{C}}{}{u^S_\psi} > 0$ represents a predefined angular velocity of considerable magnitude, enabling the ASV to execute a swift rotational motion within the time $t_1$. Additionally, $\mu$ signifies the direction of this motion, and $d^S_S$ specifies the threshold distance considered safe for the ASV's proximity to the wall. The second stage is how the AUV discerns the direction parameter $\mu$ and incorporates it into (\ref{eq:level_1_0}), without any direct communications. To address this, we define $\lambda$ as follows:
\begin{equation}\label{eq:level_2_0}
   \lambda=\frac{\prescript{\mathcal{V}}{}{\psi}^S_t}{|\prescript{\mathcal{V}}{}{\psi}^S_t|} , \, \text{when}\prescript{\mathcal{V}}{}{\psi}^S_t-\prescript{\mathcal{V}}{}{\psi}^S_{t-t_1} \geq \prescript{\mathcal{V}}{}{\psi}^S_1
\end{equation}
where $\prescript{\mathcal{V}}{}{\psi}^S_t$ denotes the yaw angle of the ASV relative to the AUV, which can be acquired using the AprilTag \cite{Wang2016} library, and $\prescript{\mathcal{V}}{}{\psi}^S_{1}$ signifies a predefined threshold. This setup facilitates the detection of rapid rotational movements of the ASV within the timeframe $t_1$. Upon detecting such movement, the AUV will modify its obstacle avoidance direction by adjusting $\lambda$, which is subsequently factored into (\ref{eq:level_1_0}) to coordinate the AUV's maneuvers.

\subsubsection{Formation Constraint}
Similar to the dog walking paradigm, the formation constraint aims to ensure a mutual connection between the two agents. In the context of visual based formation (\ref{eq:IBVS}a-b), this constraint can be represented as follows:
\begin{equation}\label{eq:formation_constraint}
\begin{aligned}
        &(W^U_t,H^U_t) \in (\Omega_S \cap \Omega_I),\\
        &(W^S_t,H^S_t) \in (\Omega_S \cap \Omega_I).
\end{aligned}
\end{equation}

The above equation indicates that, from the perspective of the image view, neither robot should deviate from the LoS at any time \(t\). Analogous to a dog leash breaking under excessive force, thereby breaching the formation constraint, similar challenges arise in multi-agent underwater systems. Efforts can be made to improve the robustness of (\ref{eq:formation_constraint}) with better IBVS parameters, but ensuring its absolute reliability is challenging due to unforeseeable disturbances. This particular challenge lies beyond the scope of our current work.

%% file: Subsections/5_experiments.tex
To validate the proposed cooperative obstacle avoidance architecture, we utilised a Gazebo-based simulation running on ROS (Robot Operating System) Melodic. 
\subsection{Simulation Setup}
\subsubsection{Robots} As introduced in Section \MakeUppercase{\romannumeral 5}, the MallARD ASV was utilised as the leader agent (walker), where the surface hydrodynamics were handled by the freefloating plugin \cite{freefloating}, while the BlueROV served as the follower agent (dog),  where the underwater hydrodynamics were managed by the UUV simulator plugin \cite{UUV}.
\subsubsection{Assumption} The underwater robot is constrained to maneuver at a fixed depth throughout the simulation. Given the cluttered environment and the reliance on visual connections between the robots, attempts to avoid obstacles by descending (\(-z\) direction) would result in the obstruction obscuring the camera's view. Conversely, ascending (\(+z\) direction) to avoid obstacles risks reducing the distance to the surface agent excessively, potentially resulting in a loss of the LoS necessary for formation.
\subsubsection{Tasks} The objectives for the two robots were to navigate from the start to the target position. The interval time threshold was set to $t_0 = 4$, and (\ref{eq:level_1_1}) was set to run at 1\,Hz in practice. The follower's initial choice for the obstacle avoidance direction was to the \textbf{left} (\(\lambda = 1\)) in all experiments. 
\subsubsection{Baseline} For comparison, we also implemented the control architecture proposed in \cite{Yao2024,YAO2023}, where the robots are connected via VET-IBVS but without implementing the dog walking paradigm.

\subsection{Case Studies on Single Obstacle Scenarios}
      \begin{figure*}[tpb]
      \centering
       \includegraphics[width=0.95\textwidth]{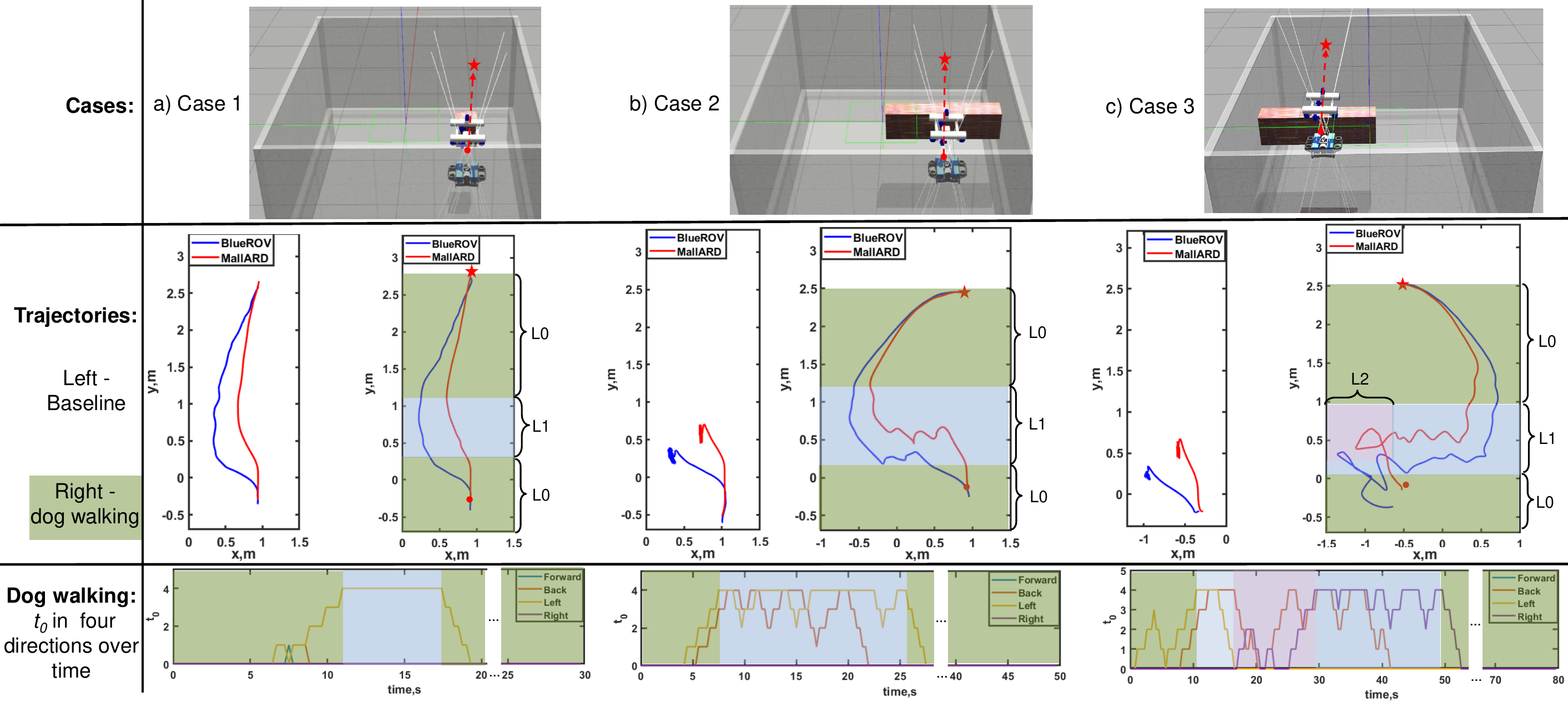}
      \caption{Comparison of case studies in single obstacle scenarios within a virtual tank measuring 5\,m x 4\,m x 2.5\,m.}
      \vspace{-15pt}
      \label{fig:all_in_one}
  \end{figure*}
\subsubsection{Case \,1 - small obstacle on the right} As depicted in Fig.~\ref{fig:all_in_one}a, MallARD leads the BlueROV from the starting point towards the target location. Upon encountering an underwater obstacle, in both baseline and dog walking systems, the BlueROV initiates a lateral maneuver to avoid collision, consequently influencing MallARD's navigation path. After circumventing the obstacle, both robots realign to the planned trajectory with the baseline and dog walking enhanced system. In dog walking system, the lateral maneuver of BlueROV results in pulling (Level\,1) MallARD to the left from \(t= 11\,s\) until they successfully circumvent the obstacle at \(t=17\,s\).

\subsubsection{Case \,2 - large obstacle on the right}As shown in Fig.~\ref{fig:all_in_one}b, upon encountering the large obstacle, in the baseline setup, MallARD, unaware of the obstacle and BlueROV's maneuver, continues toward the target. This lack of coordination causes BlueROV to become stuck, preventing both robots from reaching the target. In contrast, the "dog walking" paradigm employs three-level implicit communication and formation constraints. When BlueROV encounters the obstacle, it signals MallARD by pulling left and back, prompting MallARD to adjust its behavior and allow for successful obstacle avoidance. 

\subsubsection{Case \,3 - large obstacle on the left}


Unlike in case\,2, the system's safe direction for obstacle avoidance conflicts with the initial chosen maneuver direction (\(\lambda = 1\)), as depicted in Fig.~\ref{fig:all_in_one}c. The baseline system becomes stuck, similar to case\,2. With dog walking paradigm, BlueROV pulls MallARD left (Level\,1, blue in Fig.~\ref{fig:all_in_one}c), but as MallARD complies, it detects this path is unsafe due to proximity to the left wall. MallARD then executes a Level\,2 rotational maneuver (purple in Fig.~\ref{fig:all_in_one}c), countering BlueROV's initial pull. Recognising this cue, BlueROV adjusts its trajectory to the safe right direction (\(\lambda \rightarrow -1\)). This coordinated adjustment allows the robots to successfully avoid the obstacle, much like a walker guiding a dog away from a hazard.

\subsection{Navigation Experiment in Obscured Tank}
\begin{figure}[tpb]
      \centering
       \includegraphics[width=0.5\textwidth]{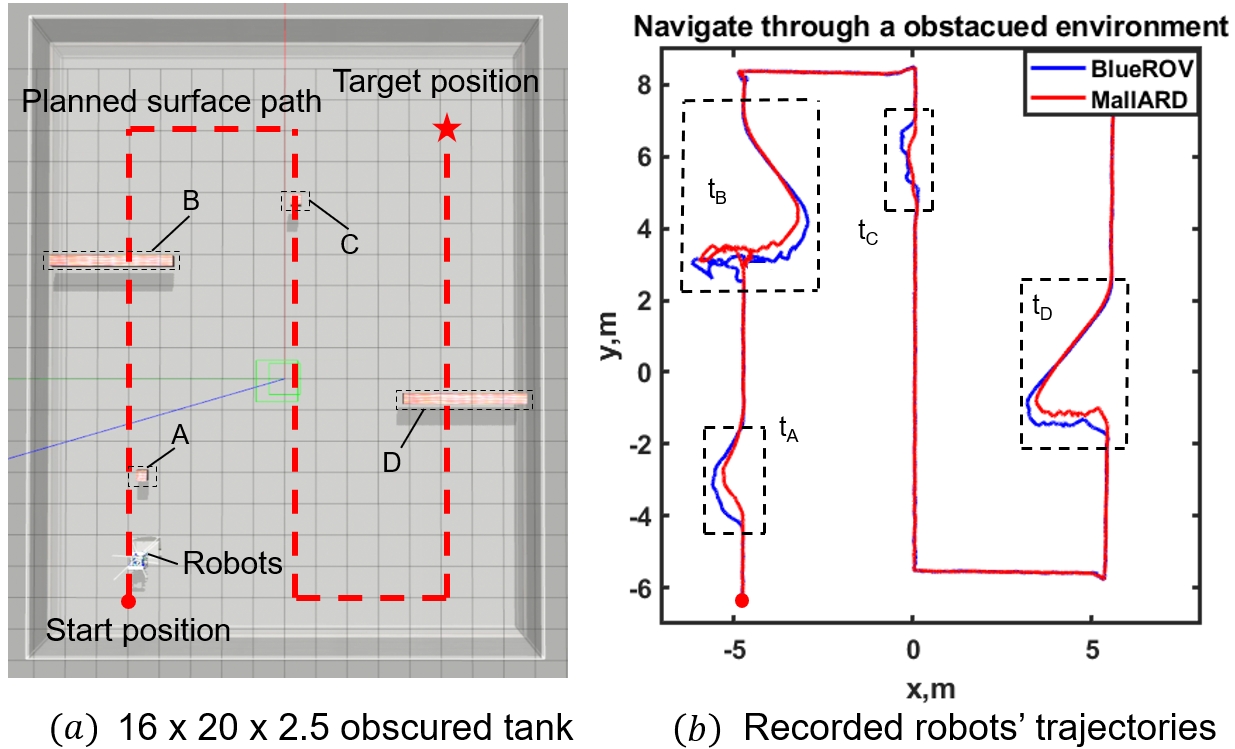}
      \caption{ Dog walking experiment: autonomous navigation in a cluttered tank with initial $\lambda = 1$. The areas $t_A$, $t_B$, $t_C$, and $t_D$ correspond to the maneuver of the robots around obstacles A, B, C, and D.}
      \vspace{-20pt}
      \label{fig:all_dog_walking}
  \end{figure}

This experiment aims to validate that, with the dog walking paradigm, the cooperative system can autonomously explore an obstacle-filled environment without collisions. As shown in Fig.\ref{fig:all_dog_walking}a, two robots are tasked to explore a confined, cluttered tank, moving from the start to the target position along a planned surface path. The environment includes various obstacles: small (A and C), large on the left (B), and large on the right (D). With an initial $\lambda = 1$, the robots are required to take actions across all levels (level 0 to level 2). The results are shown in Fig.\ref{fig:all_dog_walking}b.

\subsection{Summary}
  Results indicate that with dog walking-inspired
paradigm, fully distributed cooperative AUV-ASV underwater obstacle avoidance was achieved. It is worth noting that in previous work, authors point VET \cite{Yao2024} could potentially be useful in helping robotic obstacle avoidance in communication and localisation constrained environments, this work is the extension of VET in such direction. 

%% file: Subsections/6_conclusion.tex
This study introduces a novel approach to cooperative obstacle avoidance in cluttered underwater environments using a dog walking-inspired paradigm. By leveraging implicit communication and virtual connections like walking a dog, we demonstrate this strategy's effectiveness in coordinating multi-agent underwater systems. The analogy between dog walking and autonomous agent navigation provides a fresh perspective on distributed multi-agent systems, where role differentiation and objective balance should be emphasised. 

 The success of this solution on fully distributed cooperative underwater obstacle avoidance highlights the potential of extending the dog walking paradigm to broader scenarios across various domains (ground or aerial), particularly those characterised by extreme conditions.  While the results showcase the feasibility of our approach, the observed robot trajectories are not yet optimised. Future efforts will focus on applying advanced optimisation techniques to enhance the robustness and efficiency of the solution.